\documentclass[conference]{IEEEtran}
\usepackage{times}

\usepackage[numbers]{natbib}
\usepackage{multicol}
\usepackage[bookmarks=true]{hyperref}

\usepackage{alltt}  

\newcommand{\paperurl}{\url{bit.ly/2fnXeAd}}
\newcommand{\tuple}[1]{{\langle #1\rangle}}
\newcommand{\tup}[1]{\mbox{$\langle #1\rangle$}}

\usepackage{graphicx}
\usepackage{subcaption}

\begin{document}

\title{Combined Task and Motion Planning \\as Classical AI Planning}


\author{\authorblockN{Jonathan Ferrer-Mestres}
\authorblockA{Universitat Pompeu Fabra\\
Barcelona, Spain\\
jonathan.ferrer@upf.edu}
\and
\authorblockN{Guillem Franc\`es}
\authorblockA{Universitat Pompeu Fabra\\
Barcelona, Spain\\
guillem.frances@upf.edu}
\\
November, 2016
\and
\authorblockN{Hector Geffner}
\authorblockA{ICREA \& Universitat Pompeu Fabra\\
Barcelona, Spain\\
hector.geffner@upf.edu}
}


%

\maketitle

\begin{abstract}

Planning in robotics is often split into task and motion planning. 
The high-level, symbolic task planner decides what needs to be done, 
while the motion planner checks feasibility and fills up geometric detail.
It is known however that such a decomposition is not effective in general
as the symbolic and geometrical components are not independent. In this work,
we show that it is possible to compile task and motion planning problems
into classical AI  planning problems; i.e., planning problems over finite and
discrete state spaces with a known initial state, deterministic actions,
and goal states to be reached. The compilation is sound, meaning that classical plans are
valid robot plans, and probabilistically complete, meaning that valid robot plans are classical plans
when a sufficient number of  configurations is sampled. In this approach, motion planners and collision
checkers are used for the compilation, but not at planning time. The key elements that make the approach
effective are 1)~expressive classical AI planning languages for representing the compiled  problems in compact form,
that unlike PDDL make  use of functions and state constraints, and 2)~general width-based search algorithms capable of finding plans
over huge combinatorial spaces using weak heuristics only. Empirical results are presented for a PR2 robot manipulating tens of objects,
for which long plans are required. 
\end{abstract}

\IEEEpeerreviewmaketitle

\section{Introduction}

Planning problems in robotics involve robots that move around, while  manipulating  objects and avoiding collisions.
These problems are thought to be outside the scope of standard AI planners, and
are normally addressed through a combination of two types of planners: task planners that handle the high-level, symbolic reasoning part, and motion planners
that handle motion and geometrical constraints \cite{cambon2004asymov,gravot2005asymov,cambon2009hybrid,wolfe2010combined,lozano2014constraint,kaelbling2011hierarchical}.
These two components, however, are not independent, and hence, by giving one of the two planners  a secondary role
in the search for  plans,  approaches based on task and motion decomposition tend to be ineffective and result in lots of  backtracks \cite{lagriffoul2012constraint}. 

In recent years, there have been proposals aimed at addressing this combinatorial problem
by exploiting the efficiency of modern classical AI planners. In one case, the spatial constraints are  taken into account as part of a goal-directed replanning process where optimistic assumptions about free space  are incrementally refined until plans are obtained that can be executed in the real  environment \cite{srivastava:2014}. In another approach \cite{ffrob}, geometrical  information is used to update the heuristic used in the FF planner \cite{hoffmann:ff}. Other recent recent approaches appeal instead to SMT solvers suitable  for addressing both task planning
and the geometrical constraints \cite{nedunuri2014smt,dantam2016incremental}.

The work in this paper also aims at exploiting the efficiency of modern classical AI planning algorithms
but departs from prior work in two ways. First, task and motion  problems are  \emph{fully compiled} into  classical planning problems
so that the  classical plans are  valid robot plans. Motion planners and collision checkers \cite{lavalle:motion-planning} are used in the compilation but not in the solution of
the classical problem. The compilation is thus sound, and  probabilistically complete in the sense that robot plans map into
classical plans provided that the number of  sampled robot configurations is  sufficient.
In order to make the compiled problems compact, we move away from the standard PDDL planning language and appeal instead  to Functional STRIPS \cite{geffner:fstrips},
a planning language that is expressive enough to accommodate \emph{procedures} and \emph{state constraints}. State constraints are formulas that are forced to be true in every reachable state, and thus
represent implicit action preconditions.  In the CTMP planning encoding, state constraints are used to rule out spatial overlaps.
Procedures are used in turn for testing and updating robot and object configurations, and their planning-time execution is made efficient
by precompiling suitable  tables. The size and computation of these tables is also efficient, and 
allows us to deal with 3D scenarios involving tens of objects and a PR2 robot simulated in Gazebo \cite{koenig2004design}.

The second departure from prior work  is in the classical planning algorithm itself. Previous approaches have built upon
classical planners such as FF and LAMA \cite{hoffmann:ff,richter:lama}, yet such planners cannot be used with expressive
planning languages that feature functions and state constraints. The Functional STRIPS planner FS \cite{frances:icaps2015} handles
functions and can derive and use heuristics, yet these heuristics are expensive to compute and not always cost-effective
to deal with state constraints.  For these reasons, we build instead  on a different class of planning algorithm,
called best-first width search (BFWS), that   has  been recently shown to produce state-of-the-art results over classical planning benchmarks
\cite{nir:aaai2017}. An advantage of BFWS is that it relies  primarily on  exploratory novelty-based measures, extended with simple  goal-directed heuristics. 
For this work, we adapt  BFWS  to work with Functional STRIPS with state constraints, replacing a Functional STRIPS heuristic
that is expensive and does not take state constraints into account by a  fast and  simple heuristic suited to pick and place tasks. 

Given that classical AI planning is planning over finite and discrete state spaces with a known initial state, deterministic actions,
and a  goal state to be reached \cite{geffner:book}, it is not surprising that the combined task and motion planning can be fully compiled into a classical planning problem 
once the continuous configuration space is suitably discretized or sampled \cite{lavalle:motion-planning}. Moreover, modern classical planners
scale up very well and like SAT or SMT solvers are largely unaffected by the size of the state space. If this  approach has not been taken
before, it is thus  not due to the lack of efficiency of such planners but due to the  limitations of the languages that they support  \cite{mcdermott:pddl}.
Indeed, there is no way to compile non-overlap physical constraints into PDDL in compact form. We address this limitation by using a target language
for the compilation that makes use  of \emph{state constraints} to rule out physical overlaps  during motions, and \emph{procedures}
for testing and updating physical configurations. This additional expressive power prevents the use of standard heuristic search
planning algorithms \cite{hoffmann:ff,richter:lama} but is compatible with a more recent class of width-based planning methods that
are competitive with state-of-the-art heuristic search approaches  \cite{nir:icaps2017,nir:aaai2017}.

The paper is organized as follows. We describe first the planning language and how  the combined task and motion planning problem is modeled
as a classical problem. We present  then  the preprocessing involved, the planning algorithm, and  the  empirical results.
Videos displaying some of the problems and plans can be seen at \paperurl{}.

\section{Planning Language}

For making a general use of functions and procedures in the planning encoding, we use Functional STRIPS, 
a logical  extension of the STRIPS planning language \cite{geffner:fstrips}.  Functional STRIPS is 
a general modeling language for classical planning that is based on the variable-free fragment of first-order-logic where action $a$ have preconditions $Pre(a)$ and
effects $f(t) := t'$, where the precondition $Pre(a)$ and goals $G$, are  variable-free, first-order formulas, and $f(t)$ and $t'$ are terms with  $f$ being a fluent symbol.
Functional STRIPS  assumes that \emph{fluent} symbols, namely, those symbols whose denotation may change as a result of the actions, 
are  all \emph{function} symbols.  Constant, functional and relational (predicate) symbols whose denotation does not change 
are called  \emph{fixed} symbols, and their  denotation must be given either extensionally by enumeration, or intentionally 
by means of procedures as in \cite{dornhege2009semantic,nebel2013much}.

Terms,  atoms, and formulas  are defined  from constant, function, and relational symbols in the standard first-order-logic way, 
except that in order for the representation of states to be finite and  compact, the symbols, and hence the terms, 
are typed. A type is given by a finite set of fixed constant symbols.  The terms $f(c)$ where $f$ is a fluent symbol and $c$ 
is a tuple of fixed constant symbols are  called  \emph{state variables}, as  the state is just  the assignment of  values to
such ``variables''.

As an example, the action of moving a block $b$ onto another block $b'$ can be
expressed by an action $move(b,b')$ with precondition $[clear(b)=true  \land clear(b')=true]$, 
and effects $loc(b) := b'$ and $clear(loc(b)) := true$. In this case, the terms
$clear(b)$ and $loc(b)$ for block $b$ stand for state variables. 
$clear(loc(b))$ is a valid term, but not a state variable, as $loc(b)$
is not a fixed constant symbol. The denotation of the term $clear(loc(b))$ in a state
is a function of the $loc(b)$ and $clear(b)$ state variables; whenever
$loc(b)=b'$ holds in a state, the value of  $clear(loc(b))$  
will be that of the state variable $clear(b')$. 

Formally, a state is an assignment of values to the state variables that determines
a denotation (value) for every term and formula in the language. The denotation of a symbol or term  $t$ in the state $s$ is written as $t^s$,
while the denotation $r^s$ of terms made up of fixed symbols only and which does not depend on the state, 
is written as  $r^*$. By default, non-standard fixed constant symbols $c$, which 
usually stand for object names, are assumed to denote themselves, meaning that $c^*=c$.
The states $s$  just  encode the denotation $f^s$ of the functional fluent symbols,
which as  the types of their arguments are all finite,  can be represented as the value $[f(c)]^s$ of a finite set of state variables.
The denotation $[f(t)]^s$ of a term $f(t)$ for an arbitrary tuple of terms $t$, is then given by the value $[f(c)]^s$ of the state variable $f(c)$
where $c^* = t^s$. The denotation $e^s$ of all terms, atoms, and formulas $e$ in the state $s$ follows in the standard way.

An action $a$ is applicable in a state $s$ if  $[Pre(a)]^s = true$, and the state $s_a$ that results
from the action $a$ in $s$ satisfies the equation $f^{s_a}(t^s)=w^s$ for all the effects $f(t) := w$ of $a$, 
and otherwise is  equal to $s$. This means that the action $a$ changes the value of the \emph{state variable}  $f(c)$ to
$w^s$ in the state $s$  iff there is an effect $f(t) := w$ of action $a$ such that $t^s=c$.
For example, if $X=2$ is true in $s$,  the update $X := X + 1$ increases the value of $X$ to $3$ without affecting other state variables. 
Similarly, if $loc(b)=b'$ is true in $s$, the update $clear(loc(b)) := true$ in $s$ is equivalent to $clear(b') := true$. 

A problem  is a tuple $P=\tuple{S,I,O,G,F}$ where $S$ includes  the non-standard symbols (fixed and fluent)  and their types,
the atoms $I$ and the procedures in $F$ provide  the initial denotation $s_0$ of such symbols,  $O$ stands for the actions, and
$G$ is the goal. A plan for $P$ is  a  sequence of  applicable actions from $O$ that maps the state $s_0$ into a state $s$ that satisfies $G$.
It is assumed that standard symbols like ``$+$'', $1$, etc. have their standard denotation. Fixed functional symbols $f$ whose denotation
is given by means of procedures in $F$ are written as $@f$. The denotation of the other functional symbols must be given extensionally in $I$.

\subsection{State Constraints}

While we will make use of a  small fragment of Functional STRIPS, we will also  need a convenient extension; namely,
state constraints \cite{lin1994state,gelfond:conformant}. State constraints are formulas that are forced to be true in all reachable states,
something  achieved by interpreting state constraints as \emph{implicit action preconditions.} 
State constraints are not to be confused with state invariants that refer to formulas that are true in all reachable states
without imposing extra constraints on  actions. For example, in the blocks world, the formula
describing that no block is on two blocks at the same time is a state invariant.  On the other hand, if we assert the formula
$\neg [on(b_3,b_4) \land on(b_4,b_5)]$ as a state constraint, we are ruling out actions leading to states where 
the formula $[on(b_3,b_4) \land on(b_4,b_5)]$ holds. 

A Functional STRIPS  problem with \emph{state constraints} is a tuple $P'= \tuple{S,I,O,G,C,F}$ where the new component $C$ stands
for a set of formulas expressing the state constraints. The syntax for these formulas is the same as for those encoding (explicit)
action preconditions but their semantics is  different: an action $a$ is deemed applicable in a state $s$  when \emph{both} $[Pre(a)]^s = true$ \emph{and}
the state $s_a$ that results from applying $a$ to $s$ is such that $c^{s_a} = true$ for every state constraint $c \in C$. 

A plan for $P'$ is thus  a  sequence of  actions  from $O$ that maps the state $s_0$ into a state $s$ that satisfies $G$,
and such that for each such action $a$, $Pre(a)$ is true in the state $s$ where the action $s$ is applied, and all constraints in $C$ are true in the resulting state. 
It is assumed  that the state constraints hold in the initial state.

\section{Modeling Pick-and-Place Problems}

\begin{figure}[t] 
\footnotesize
\centering
\begin{alltt}
(:action MoveBase
 :parameters (?e - base-graph-traj-id)
 :prec (and (= Arm ca0)
            (= Base (@source-b ?e))
 :eff (and (:= Base (@target-b ?e))))

(:action MoveArm
 :parameters (?t - arm-graph-traj-id)
 :prec (and (= Arm (@source-a ?t))
 :eff (and (:= Arm (@target-a ?t))
           (:= Traj  ?t)))

(:action Grasp
 :parameters (?o - object-id)
 :prec (and  (=  Hold  None)   
  (@graspable Base Arm (Conf ?o)))
 :eff (and (:= Hold ?o)
           (:= (Conf ?o) c-held)))

(:action Place
 :parameters (?o - object-id)
 :prec  (and (= Hold ?o)
    (@placeable Base Arm)
 :eff (and (:= Hold None)
   (:= (Conf ?o)(@place Base Arm)))

(:state-constraint
  :parameter (?o - object-id)
   (@non-overlap Base Traj (Conf ?o) Hold))
 
\end{alltt}
\caption{\footnotesize  CTMP Model Fragment in Functional STRIPS: Action and state constraint schemas. Abbreviations used. Symbols preceded by ``@'' denote
  procedures.  All objects  assumed to have the same shape. Initial situation provides initial values for the state variables \emph{Base}, \emph{Arm}  (resting), $Traj$ (dummy), and
$Conf(o)$ for each object. Goals describe target object configurations. State constraints prevent collisions during arm motions.
Motion planners and collision checkers used  at compilation time, not at plan time, as detailed in the Preprocessing section.}
\label{fig:modeling}
\end{figure}

We consider CTMP problems involving a robot and a number of objects located on tables of the same
height. The  tasks involve   moving some objects from some initial configuration to a final configuration or set of configurations,
which may require moving obstructing  objects as well. The model is tailored to a PR2 robot using  a single arm, but
can be generalized easily.

The main  state variables \emph{Base}, \emph{Arm}, and \emph{Hold} denote the configuration of the robot base,
the arm configuration, and the content of the gripper, if any. In addition, for each object $o$, the
state variable \emph{Conf(o)} denotes  the configuration of object $o$. The configuration of the robot base represents the 2D position
of the base and its orientation angle. The configuration of the robot arm represents the configuration of the end effector:
its 3D position, pitch, roll, and yaw. Finally, object configurations are 3D positions, as for simplicity we consider object
that are symmetric, and hence their orientation angle is not relevant. There is also a state variable $Traj$,
encoding the last trajectory followed by the robot arm, which is needed  for checking collisions during arm motions. 
All configurations and trajectories are obtained from a \emph{preprocessing stage}, described in the next section,
and are represented in the planning encoding by symbolic ids. When plans are executed, trajectory ids become motion plans;
i.e.  precompiled sequences  of base and arm join vectors, not represented explicitly in the planning problem.

The encoding assumes  \emph{two finite graphs}: a \emph{base graph},  where the nodes stand for  robot base configurations
and  edges stand for trajectories among pairs of base  configurations, and an \emph{arm graph},  where  nodes stand for
end-effector configurations (relative to a fixed base), and edges  stand for arm trajectories  among pairs of such configurations.
The details for how such graphs are generated are not relevant for the planning encoding and will be described below.
As a reference, we will consider instances with tens of objects, and base and arm graphs with hundreds of
configurations each, representing thousands of robot configurations. 

A fragment of the planning encoding featuring all the actions and the state constraints is shown in Figure~\ref{fig:modeling}.
Actions $MoveBase(e)$ take an edge $e$ from the base graph as an argument, 
and update the base configuration of the robot to the target configuration associated
with the edge. The precondition is that the  source configuration of the edge corresponds
to the current base configuration, and that the arm is the resting configuration \emph{ca0}.
Actions $MoveArm(t)$ work in the same way, but the edges $t$ of the arm graph are used instead. 


There are also  actions \emph{Grasp(o)} and \emph{Place(o)}  for grasping and placing objects $o$. The grasping action
requires that the gripper is empty and that \emph{@graspable(Base,Arm,Conf(o))} is true,  where the  procedure
denoted by the symbol \emph{@graspable}  checks if  the robot configuration,  as determined by its base and (relative) arm  configuration,
is such that object $o$ in its current configuration can be grasped by just closing the gripper.
Likewise, the atoms $Hold=o$ and  \emph{@placeable(Base,Arm,Conf(o))} are  preconditions of the action $Place(o)$ .

The total number of ground actions is given by the \emph{sum} of the number of edges in the two graphs and the number of objects. 
This small number of actions is made possible by the  planning language  where robot, arm, and object configurations
do not appear as action arguments. The opposite would be true in  a STRIPS encoding where action
effects are determined solely by the action (add and delete lists) and do not depend on the state.
The number of state variables  is also small, namely, one state variable for each object and four other state variables. 
Atoms whose predicate symbols denote procedures, like \emph{@graspable(Base,Arm,Conf(o))}, do not represent state variables or fluents, 
as  the denotation of such predicates is fixed and constant. These procedures play a key role
in the encoding, and in the next section we look at the preprocessing that converts
such procedures into fast lookup operations.

The only subtle aspect of the encoding is in the state constraints used to prevent collisions. 
Collisions  are  to be avoided not just at beginning and end of actions, but also during  action
execution. For simplicity,  we assume that robot-base moves do not cause collisions
(with mobile objects), and hence that collisions result exclusively from arm motions. We enforce this
by restricting the mobile objects to be on top of tables that are fixed, and by requiring the arm to
be in a suitable resting configuration  (ca0)  when the robot base moves.
There is one state constraint  \emph{@nonoverlap(Base,Traj,Conf(o),Hold)} for each object $o$, 
 where $Traj$ is the state variable  that keeps track of the last arm trajectory
executed by the robot. The procedure denoted by the symbol \emph{@nonoverlap} tests whether a collision occurs between
object $o$ in configuration \emph{Conf(o)} when  the robot arm moves  along the trajectory $Traj$ and the robot base configuration
is \emph{Base}. The test depends also on whether the gripper is holding an object or not. As we will show in the
next section, this procedure is also  computed  from  two \emph{overlap  tables}
that are precompiled  by calling the MoveIt collision-checker \cite{moveit} a number of times that is twice the
number of edges (trajectories) in the arm graph. 


\section{Preprocessing}

The planning  encoding shown in Fig.~\ref{fig:modeling} assumes a crucial preprocessing stage where
the base and arm graphs are computed, and suitable tables are stored for avoiding the use
of motion planners and collision checkers during planning time. This  preprocessing is efficient and 
does not depend on the number of objects, meaning it can be used for several problem
variations without having to call collision checkers and motion planners again. Indeed,
except for the overlap tables, the rest of the compilation is local and does not depend
on the possible robot base configurations at all. 

To achieve this, we consider the robot at a \emph{virtual base } $B_0 = \tup{x,y,\theta}$
with $x=y=\theta=0$ in front of a \emph{virtual table} whose height is the height of the actual tables,
and whose dimensions exceed the (local) space that the robot can reach without moving the base. 
By considering the robot acting in this local virtual space without moving from this virtual base $B_0$,  we will obtain
all the relevant information about object configurations and arm trajectories, that  will carry to the real robot base configurations $B$ 
through a simple linear transformations that depend on $B$. The computation of the overlap tables is more subtle and will be considered later.

First of all, the $x,y$ space of the virtual table is discretized regularly into
$D$ position pairs $x_i,y_i$. If the height of the objects is $h'$ and the height of the tables is $h$, 
then the   virtual object configurations are set to the triplets $\tup{x_i,y_i,z}$ where $z=h+h'/2$.
Each virtual object configuration  represents a  possible center of mass for the objects when sitting at location $x_i,y_i$
over the virtual table. For each such configuration  $C = \tup{x_i,y_i,z}$,  $k$ grasping poses $A_C^j$ are defined  from which an  object
at  $\tup{x_i,y_i,z}$ could be grasped, and a motion planner (MoveIt) is called to compute $k'$ arm trajectories
for reaching each such grasping pose $A_c^j$ through $k'$ different waypoints from a fixed resting pose and
the robot base fixed at $B_0$. This means that up to $k \times k'$ arm trajectories are computed for each virtual object configuration,
resulting in up to $D \times k \times k'$ arm trajectories in total and up to $k \times D$ grasping poses.
For each reachable grasping pose $A_C^j$, we store the pair $\tup{A_C^j,C}$ in a hash table. The table
captures the function $vplace$ that   maps grasping poses (called arm configurations here),
into virtual object configurations. The meaning of $vplace(A)=C$ is that when the robot base is at $B_0$ in front the virtual table
and the arm configuration is $A$, an object on the gripper will be placed at the virtual object configuration  $C$.

The \emph{arm graph} has as nodes the arm configurations $A$ that represent reachable grasping poses $A=A_C^j$ in relation to
some virtual object configuration $C$, in addition to the resting arm configuration. The arm trajectories that connect
the resting arm configuration $A_0$  with an arm trajectory $A$ provide the edge in the arm graph between $A_0$ and $A$.
The graph contains also the inverse edges that correspond to the same trajectories reversed. 
Grasping configurations  that are not reachable with any trajectory from the resting arm configuration are pruned
and virtual object configurations all of whose grasping poses have been pruned, are pruned as well.

The  \emph{base graph} is computed by sampling a number of configurations $N_B$ near the tables
and calling the MoveIt motion planner to connect each such configuration with  up to $k_B$ of its closest neighbours.
The number of \emph{robot configurations} results from the product of the number of arm configurations $k \times D$
and the number of base configurations $N_B$. In the experiments we consider numbers that go from tens to a few hundred
and which thus result into  thousands of possible robot configurations.  The computation of the base and 
arm graphs defines the procedures used in the $MoveBase$ and $MoveArm$ actions that access the source and
target configuration of each graph edge.

The set of (real) \emph{object configurations}  are then  defined and computed  as follows. The virtual object configuration
$C=\tup{x,y,z}$ represents the 3D  position of the object before a pick up or after a place action, 
with the arm at configuration $A$ and the robot base at the virtual base configuration  $B_0=\tup{0,0,0}$.
As the robot moves from this ``virtual'' base  to an arbitrary  base $B$ in the base graph, the point $C$
determined by the same arm configuration $A$ moves to a new point $C'$ that is given by a transformation
$T_B(C)$ of $C$ that depends solely on $B$. Indeed, if $B = B_0 + \tup{\Delta_X, \Delta_Y,\Delta_{\theta}}$ with $\Delta_{\theta}=0$, 
then $T_B(C) = \tup{x+\Delta_X,y+\Delta_Y,z}$. More generally, for any $\Delta_{\theta}$,  $T_B(C)  = \tup{x',y',z}$ with $x' = \Delta_X + (x - \Delta_X)cos(\Delta_{\theta}) - (y - \Delta_Y)sin(\Delta_{\theta})$ and $y' = \Delta_Y + (x - \Delta_X)sin(\Delta_{\theta}) + (y-\Delta_Y)cos(\Delta_{\theta})$.
The set of \emph{actual object configurations} is then given by such  triplets $T_B(C)=\tup{x',y',z}$
for which 1)~$B$ is a  node of the base graph, 2)~$C$ is a virtual object configuration, and 3)~the 2D point $x',y'$ falls within
a table in the actual environment. That is, while the virtual object configurations live only in the virtual table with the base fixed at $B_0$,
the actual object configurations depend on the virtual object configurations, the base configurations, and the real tables in the working space.
We will write $T_B(C)=\bot$ when $C$ and $B$ are such that for  $T_B(C)=\tup{x',y',z'}$, the 2D point $x',y'$ does not fall within
a table in the actual environment. In such a case,  $T_B(C)$ doesn't denote an actual object configuration.

Given the  linear transformation $T_B$ and the function $vpose(A)$ defined above, that maps an arm  configuration  into a  virtual object configuration
that is relative to the virtual base $B_0$, the procedures denoted by the symbols $@graspable$, $@placeable$, and $@pose$ in the planning encoding are defined as follows:

\begin{eqnarray*}
@pose(B,A)  =  C' &&  \hbox{iff} \  C'= T_B(vpose(A)) \\
@graspable(B,A,C')  =  true  &&  \hbox{iff} \  C'= @pose(B,A) \\
@placeable(B,A)   =  true  &&  \hbox{iff} \ @pose(B,A) \not= \bot \ . 
\end{eqnarray*}

\medskip

We are left to specify the compilation of the tables required for computing the 
$@nonoverlap$ procedure without calling a collision checker at planning time.
This procedure  is used in the  state constraints \emph{@nonoverlap(B,Traj,Conf(o),Hold))}
for ruling out actions that move the arm along a  trajectory $Traj$ such that for the current base configuration $B$
and content of the gripper $Hold$, will cause a collision with some object $o$ in its current configuration $Conf(o)$. 
For doing these tests at planning time efficient, we precompile two additional tables, called the holding and non-holding overlap tables
(HT, NT), which  are made of pairs $\tup{Tr,C}$ where $Tr$ is a trajectory in the arm graph, and $C$ is what we will call a \emph{relative object configuration}
different than both the virtual and real object configuration. Indeed, the set of relative object configurations is defined as the set of configurations
$T^{-1}_B(C)$ for all bases $B$ and all real object configurations $C$, where $T_B^{-1}$ is the inverse of the linear transformation $T_B$ above.
If $C$ is a real 3D point obtained by mapping a point $C'$ in the virtual table after the robot base changes from $B_0$ to $B$,
then $C''=T_{B'}(C)$ for $B'=B$ is just $C'$ but for $B'\not= B$, it  denotes a point in the ``virtual'' space relative to the base $B_0$
that may not  correspond to a virtual object configuration, and may even fail to be in the space of the virtual table
(the local space of the robot when fixed at base $B_0$).  Relative  object configurations $C''$ that do not fall within the virtual table,
are pruned. The \emph{holding overlap table (HT)}  contains then the pair $\tup{Tr,C}$  for a  trajectory $Tr$
and a relative object configuration,  iff the robot arm moving along trajectory $Tr$ will collide with an object in the virtual configuration $C$
when the robot base is at $B_0$ and the gripper is carrying an object. Similarly, the pair  $\tup{Tr,C}$ belongs to the \emph{non-holding overlap table (NT)}
iff the same condition arises when the gripper is empty. Interestingly, each of these two tables is compiled by calling a collision checker (MoveIt)
a number of times that is given by the total number of arm trajectories.  Indeed, for each trajectory $T$,  the collision checker tests in one single scan  which
relative configurations $C$ are on the way.

The procedure \emph{@nonoverlap(B,Tr,Conf(o),Hold)} checks whether trajectory $Tr$ collides with object $o$ in configuration $Conf(o)$
when the robot base is $B$. If $Hold$ is $None$, this is  checked by testing whether the pair $\tup{Tr,T_B^{-1}(Conf(o))}$ is in the NT table,
and if $Hold$ is not $None$, by testing whether the pair is in the HT table.  These are   lookup operations in the two (hash) tables NT and HT,
whose size is determined by the number of trajectories and the number of relative object configurations. This last number is independent
of the number of objects but higher than the number of virtual configurations. In the worst case, it is bounded by
the product of the number $N_B$ of robot bases and the number of real object configurations, which in turn is bounded by $N_B \times N_C$,
where $N_C$ is the number of virtual object configurations.  Usually, however, the number of entries in the overlap tables NT and HT
is much less, as for most real object configurations $C$ and base $B$, the point $T^{-1}_B(C)$ does not fall into the ``virtual table''
that defines the local space of the robot when fixed at $B_0$. The size of the hash table $\tup{Tr,C}$ precompiled for encoding 
the function $vpose(Tr)$ above is smaller and given just by the number of arm trajectories $Tr$, 
to the number of edges in the arm graph, which in turn is equal to  $2 \times D \times k \times k'$, where
$D$ is the number of virtual object configurations, $k$ is the number of grasping poses for each virtual object configuration, 
and $k'$ in the number of trajectories for reaching each grasping pose.

\section{Planning Algorithm}

The compilation of task and motion planning problems  is efficient and results in planning problems that are compact.
Yet, on the one hand, standard planners like  FF and LAMA  do not handle functions and state constraints,
while planners that do compute heuristics that in this setting are not cost-effective
\cite{frances:icaps2015}.  For these reasons, we build instead  on a different class of planning algorithm, called best-first width search (BFWS),
that  combines some of the benefits of the goal-directed heuristic search with those of width-based search
\cite{nir:ecai2012}. 

Pure width-based search algorithms are exploration algorithms and do not
rely on goal directed heuristics. The simplest such algorithm  is IW(1), which is a
plain breadth-first search where newly generated  states that do not make an atom $X=x$ true for
the first time in the search are pruned. The algorithm IW(2) is similar except that a state $s$ is pruned
when there are no atoms $X=x$ and $Y=y$ such that the \emph{pair} of atoms $\tup{X=x,Y=y}$
is true in $s$ and false in all the states generated before $s$. More generally,
the algorithm IW($k$) is a normal breadth-first except that newly generated
states $s$ are pruned when their ``novelty'' is greater than $k$,
where the \emph{novelty}  of $s$ is $i$ iff there is a tuple $t$ of $i$
atoms such that $s$ is the first state in the search that makes all the atoms
in $t$ true, with  no tuple of smaller size having this property \cite{nir:ecai2012}.
While simple, it has been shown that the procedure  IW($k$) manages
to  solve arbitrary instances of many of the standard benchmark domains 
in low polynomial time  provided that the goal is a single atom. 
Such domains can be shown indeed to have a small and bounded \emph{width} $w$ that
does not depend on the instance size, which implies that they can be
solved (optimally) by running IW($k$). Moreover, IW($k$) runs in time
and space that are exponential in $k$ and not in the number of problem
variables. 
IW calls the procedures IW(1), IW(2), \ldots\
sequentially until finding a solution. IW is complete but not effective in problem
with multiple goal atoms. For this, \emph{Serialized IW} (SIW) calls IW sequentially
for achieving the goal atoms one at a time.  While SIW is a blind search procedure that is incomplete (it can get trapped into dead-ends),
it turns out to  perform much better than  a greedy best-first guided by the standard heuristics. 
Other variations of IW have  been used for planning   in the Atari games and those of the General Video-Game AI competition
\cite{nir:ijcai2015,tomas:aiide2015,carmel:ijcai2016}.

Width-based algorithms such as IW and SIW do not require PDDL-like planning models and can work directly with simulators, and
thus unlike heuristic search planning algorithms, can be easily adapted to work  with Functional STRIPS with state constraints.
The problem is that by themselves,  IW and SIW  are  not  powerful enough for solving large CTMP problems.
For such problems it is necessary to complement the effective exploration that comes from 
width-based search  with the guidance that results  from goal-directed heuristics. 
For this reason, we appeal to a combination of heuristic and width-based search called
Best-First Width Search (BFWS), that has been shown recently to yield state-of-the-art  results
over the classical planning benchmarks \cite{nir:aaai2017}. BFWS is a standard best-first
search with a sequence of evaluation functions $f=\tup{h,h_1, \ldots, h_n}$  where 
the node that is selected for expansion from the OPEN list at each  iteration 
is the node that minimizes $h$, using the other $h_i$ functions lexicographically
for breaking ties. In the best performing variants of BFWS, the main  function $h=w$
computes the ``novelty''  of the nodes, while the other functions $h_i$
take the goal into account. 

For our compiled CTMP domain, we use BFWS with an  evaluation function $f=\tup{w,h_1,\ldots, h_n}$,
where $w$ stands for a standard novelty measure, and  $h_1, \ldots, h_n$ are simple heuristic counters defined
for this particular domain. The novelty $w$ is defined as in \cite{nir:aaai2017}; namely,  the novelty $w(s)$  of a newly generated state $s$ in the BFWS guided
by the function  $f= \tup{w,h_1, \ldots, h_n}$ is $i$ iff there is a tuple (conjunction)  of $i$ atoms $X=x$, and no  tuple of smaller size,  that is  true in $s$ but false
in all the  states $s'$ generated before $s$ with the same function values $h_1(s')=h_1(s)$, \ldots, and $h_n(s')=h_n(s)$.
According to this definition, for example, a new state $s$ has novelty 1 if there is an atom $X=x$
that is true in $s$ but false in all the  states $s'$ generated before $s$  where $h_i(s')=h_i(s)$ for all $i$. 

For the tie-breaking functions $h_i$ we consider three counters.  The first is the standard  goal counter
$\#g$ where $\#g(s)$ stands for  the number of goal atoms that are not true in $s$. 
The second is an slightly richer goal counter $h_{M}$ that takes into account that each object that has to  be moved to a
goal  destination has to involve  two actions at least:  one for picking up the object, and one for placing the object.
Thus $h_M(s)$  stands for twice the number of objects that are not in their  goal configurations in $s$, minus $1$
in case that one such object is being held.

\begin{figure*}[!tbp]
  \begin{subfigure}[b]{0.45\textwidth}
    \includegraphics[width=\textwidth]{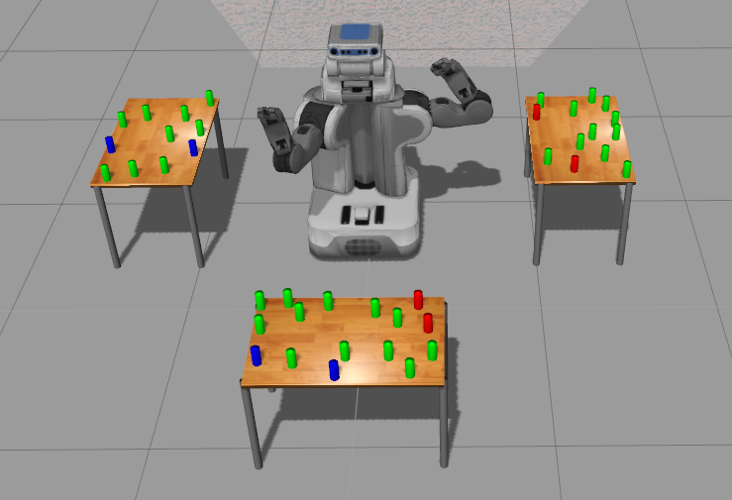}
  \end{subfigure}
  \hfill
  \begin{subfigure}[b]{0.45\textwidth}
    \includegraphics[width=\textwidth]{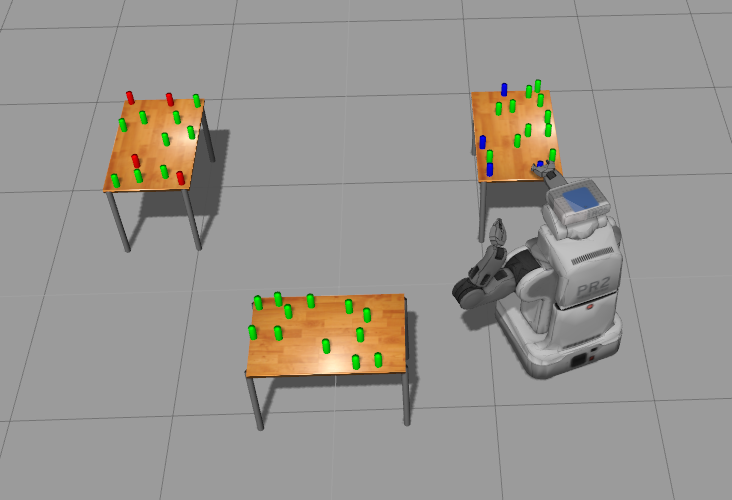}
  \end{subfigure}
  \caption{Manipulating objects in a 3-table environment, initial (left) and goal (right) situations. The objective is to put the blue objects on the rightmost table and the red objects on the leftmost table.}
  \label{fig:images}
\end{figure*}

The last tie-breaker used corresponds to the counter $\#c(s)$ that tracks  the number of objects that are in ``obstructing configurations'' in the state $s$.
This measure is determined from a set $\cal C$ of object configurations  $C$  computed once from the initial problem state, as it is common
in landmark heuristics.  The count $\#c(s)$ is $i$ if there are $i$ objects $o$ for which the state variable $Conf(o)$ has a value in $s$ that is 
in $\cal C$. The intuition is that a configuration is ``obstructing'' if it's on the way of an  arm trajectory  that follows  a suitable  relaxed plan
for achieving a  goal atom. More precisely, we use a single IW(2) call  at preprocessing for computing a   plan  for each goal atom
in a problem relaxation that ignores state constraints (i.e., collisions). These relaxed problems are ``easy'' as they just involve robot motions
to pick up the goal  object followed by a  pick up action,  more robot motions, and  a place action. The search tree constructed by IW(2) normally includes 
a plan for each goal atom in this relaxation, and often more than one plan. One such \emph{relaxed plan} ``collides'' with an object $o$ if a $MoveArm(t)$ action
in the plan leads to a state where a state constraint \emph{@nonoverlap(Base,Arm,Conf(o),Hold)} is violated (this is possible
because of the relaxation). In the presence of multiple plans for an atomic goal in the relaxation, a plan is selected
that collides with a minimum number of objects. For such an atomic goal, the ``obstructing configurations'' are the 
real object configurations $C$ such that a state constraint \emph{@nonoverlap(Base,Arm,C,Hold)} is violated in some state of the relaxed plan
where $Conf(o)=C$ for some object $o$. We further consider as obstructing those configurations that 
in a similar manner obstruct the achievement of the goal of holding any object $o$ that is in an obstructing configuration
\emph{in the initial state}, recursively and up to a fixpoint.
The set $\cal C$ is then the union of the sets of ``obstructing configurations'' for each atomic goal,
and $\#c(s)$ is the number of objects $o$ for which  the value $C$ of the state variable  $Conf(o)$ in $s$ belongs to $\cal C$.
Note that unlike the other two heuristics $\#g$ and $h_M$, which must have  value zero in the goal, the  $\#c(s)$ counter
may be different than zero in the goal. Indeed, if a  problem  involves exchanging the configuration  of two objects,
$\#c(s)$ will be equal to $2$ in the goal, as the two goal configurations are actually obstructing configurations
as determined from the initial state. The set $\cal C$ of obstructing configurations is computed once from the
initial state in low polynomial time by calling the IW(2) procedure once. The resulting $\#c(s)$ count
provides an heuristic estimate of the number of objects that need to be removed in order to achieve the goal,
a version of the minimum constraint removal problem \cite{min-removal} mentioned in \cite{ffrob}.

The counters  $h_M$ and $\#c$ used in the BFWS algorithm for CTMP planning can be justified on domain-independent grounds. 
Indeed, $h_M$  corresponds roughly to the cost of  a problem where both state constraints and  preconditions involving procedures have been relaxed.
So the plans for the relaxation are sequences of pickup and place actions involving the goal objects only. The counter $\#c$ is related to landmark heuristics
under the assumption that the goals will be achieved through certain motion plans.

The  third element  in our BFWS algorithm  is the extension of the problem states with two extra Boolean features
\emph{graspable*(o)} and \emph{placeable*(o)} associated with each object $o$.
The features  \emph{graspable*(o)} and \emph{placeable*(o)} are set to true in a state $s$ iff the preconditions of the actions \emph{Grasp(o)} and \emph{Place(o)}
are true in $s$ respectively. These features are needed  as there are no  state variables  related to
the preconditions \emph{(@graspable B A Conf(o))} and  \emph{(@placeable B A)} of those actions, as  the predicate symbols of these atoms denote procedures.
That is, the terms $B$, $A$, and $Conf(o)$  in these atoms denote state variables but the relations themselves, denoted by the symbols \emph{@graspable} and \emph{@placeable},
are static. 

Finally, for the experimental results we have found useful to add an extra precondition to the action $MoveArm(t)$. This precondition
requires that \emph{@target-a(t)} is the resting configuration \emph{ca0} or that \emph{@placeable(Base,@target-a(t))} is true. 
In other words, the arm is moved from the resting position to configurations where an object could be picked up or placed.
This restriction reduces the average branching factor of the planning problem, in particular when the number of arm motions in the
arm graph is large. 

\section{Experimental Evaluation}

\begin{table*}[t]
  {
    \small
   \begin{center}
    \begin{tabular}{rrrrrrrrrr} 
    
    \hline 
 tables & trajectories & arm conf. & base conf. & total conf. & virtual conf. & virtual GP & relative conf. & real conf. &  Time(min.) \\ \hline 
  1 & 268 & 43 & 124 & 5332 & 15 & 42 & 1081 & 136 & 5\\
  3 & 268 & 43 & 323 & 13889 & 15 & 42 & 3379 & 393 & 13\\
    \hline 
    \end{tabular}
  \end{center}
  }
\caption{\small 
\emph{Compilation data for one and three tables}.
Columns show the number of tables, total number of arm trajectories, arms configurations, base configurations, total number of robot configurations, virtual object configurations, number of virtual grasping poses, relative object configurations, total number of real object configurations and overall compilation time.
  } \label{tab:compilation}     
\end{table*}

\begin{table*}[t]
  {
     \small

\begin{subtable}{.5\linewidth}\centering
{\begin{tabular}{rrrrrrrrr} 
    
    \hline 
 \#o  &  \#g  & \#c  & & L & E & Prep  & Search  &  Total \\ \hline 

  10 & 1 & 4  &  &  38 & 700 & 2.4 & 0.08 & 2.48\\
  
  10 & 2 & 6  &  &  67 & 5.7k & 2.42 & 0.64 & 3.06\\
  
  10 & 3 & 8 &  &  73 & 6.1k & 2.22 & 0.72 & 2.94\\

  15 & 1 & 6  &  &  49 & 778 & 3.4 & 0.1 & 3.5\\

  15 & 2 & 8  &  &  81 & 9.8k & 3.76 & 1.27 & 5.03\\

  15 & 3 & 10 &  &  80 & 7.7k & 4.13 & 0.97 & 5.1\\

  20 & 1 & 12 &  &  86 & 39k & 5.44 & 4.46 & 9.9\\

  20 & 2 & 14 &  &  122 & 63.3k & 5.85 & 9.42 & 15.27\\

  20 & 3 & 22 &  &  159 & 49.2k & 5.66 & 7.26 & 12.92\\

  25 & 1 & 4 &  &  22 & 206 & 7.42 & 0.03 & 7.45\\ 

  25 & 2 & 4 &  &  45 & 39.1k & 7.29 & 5.54 & 12.83\\

  25 & 3 & 18 &  & \texttt{MO} & - & - & - & -\\
  
  30 & 1 & 4 &  &  22 & 67.6k & 9.21 & 10.16 & 19.37\\

  30 & 2 & 38 &  &  \texttt{MO} & - & - & - & - \\

  30 & 3 & 38 &  &  \texttt{TO} & - & - & - & -\\
    \hline 
    \end{tabular}}
\caption{Manipulating objects, one single table.}\label{tab:results-1}
\end{subtable}%
\begin{subtable}{.5\linewidth}\centering
{\begin{tabular}{rrrrrrrrr} 
    \hline 
 \#o  &  \#g  & \#c  & & L & E & Prep & Search  &  Total \\ \hline 

  10 & 2 & 6 &  &  54 & 1.3k & 8.1 & 0.23 & 8.33\\
  10 & 4 & 2 &  &  101 & 3.9k & 8.1  & 0.8 & 8.9\\
  
  10 & 6 & 2 &  &  121 & 3.9k & 7.18 & 0.6 & 7.78\\

  10 & 8 & 2 &  &  150 & 4.5k & 8.26 & 0.91 & 9.17\\

  20 & 2 & 4 &  &  65 & 6.2k & 19.19 & 1.32 & 20.51\\

  20 & 4 & 4 &  &  89 & 9.6k & 17.9 & 2.29 & 20.19\\

  20 & 6 & 6 &  &  130 & 3.1k & 17.66 & 0.73 & 18.39\\

  20 & 8 & 8 &  &  141 & 5.9k & 18.42 & 1.26 & 19.68\\

  25 & 2 & 8 &  &  46 & 1.1k & 23.74 & 0.23 & 23.97\\

  25 & 4 & 8 &  &  80 & 2.3 & 24.44 & 0.54 & 24.98\\

  25 & 6 & 10 &  &  120 & 3.5k & 27.04 & 0.91 & 27.95\\

  25 & 8 & 12 &  &  158 & 3.4k & 23.74 & 0.69 & 24.43\\
  
  30 & 2 & 4 &  &  \texttt{MO} &-  & - & - & - \\

  30 & 4 & 2 &  &  74 & 1.6k & 30.37 & 0.4 & 30.77\\

  30 & 6 & 8 &  &  123 & 2.6k & 30.09 & 0.64 & 30.73\\
  
  30 & 8 & 10 &  &  161 & 3.5k & 32.22 & 0.86 & 33.08\\

  40 & 2 & 4 &  &  52 & 1.3k & 45.64 & 0.33 & 45.97\\

  40 & 4 & 14 &  &  114 & 55.5k & 45.65 & 13.12 & 58.77\\

  40 & 6 & 10 &  &  178 & 166k & 47 & 41.36 & 88.36\\

  40 & 8 & 14 &  &  220 & 201k & 46.46 & 55.57 & 102.03\\
    \hline 
\end{tabular}}
\caption{Manipulating objects, three tables.}\label{tab:results-3}
\end{subtable}
  }
\caption{\small 
\emph{{Per-instance results} for one and three tables}.
Each row shows results for one instance. Three leftmost columns report instance characteristics. \emph{\#o} denotes number of objects on the table,
\emph{\#g} number of different goals, \emph{\#c} is a proxy for the number of objects 
that initially obstruct the achievement of the goal, as described in the text. Remaining columns report length of the computed plan (\emph{L}),
number of nodes expanded during the search (\emph{E}), and, in seconds,
preprocessing, search and total time. \texttt{TO} and \texttt{MO} denote time- and memory-outs.
} 
\label{tab:results-all}
\end{table*}

We test our model on two environments having one and three tables,
the characteristics of which are shown in Table~\ref{tab:compilation}.
As explained above, the virtual space of the robot is discretized into $D=15$ position pairs or \emph{virtual configurations},
with $k=4$ grasping poses per virtual configuration and $k'=4$ arm trajectories for each
of those grasping poses, obtained from Moveit. Thus, the maximum number of (virtual) grasping poses will
be $D \times k = 60$, of which those for which no motion plan is found get pruned. In our benchmark environments,
the total number of virtual grasping poses is $42$.
In turn, the maximum number of arm trajectories is $D \times k \times k' = 240$ in each direction,
i.e. $480$, while in both of our environments we have a total of $268$ such trajectories, since again no feasible
motion plans are found for the rest.
The number of sampled bases is $124$ for the one-table environment and $323$ for the three-table environment,
while each robot base in the base graph is connected to a maximum of $12$ closest base configurations.
Importantly, the output of the precompilation phase, which takes 5 min. (13 min.) for the one-table (three-tables) environment,
is valid for for all instances with that number of tables, regardless of number of objects, initial robot and object configurations,
and particular goals of the problem.

For each environment, we generate a number of semi-random instances with increasing number of
objects, ranging from $10$ to $40$, and increasing number of goals, ranging from $2$ to $8$, where
a problem with e.g. $4$ goals might require that $4$ different objects be placed
in their respective, given target configurations.
The initial and goal states of a sample problem instance are shown in Fig.~\ref{fig:images},
where the robot needs to place all blue objects in one table and all red objects in another.
Tables~\ref{tab:results-1} and \ref{tab:results-3} show the results of our BWFS planner on each generated instance,
running with a maximum of 30 minutes and 8GB of memory on an AMD Opteron 6300@2.4Ghz. 
The planner uses ROS \cite{quigley2009ros}, Gazebo \cite{koenig2004design}, and MoveIt \cite{moveit},
in the preprocessing and in the simulations, but not at planning time.
Videos showing the execution of the computed plans in the Gazebo simulator, for some selected instances,
can be found in \paperurl{}.
The results show that our approach is competitive and scales well with the number of objects in the table.
The length of the obtained plans ranges from $22$ to $220$ steps.
Problems with up to $20$ objects, both for one and three tables, for example, are solved in a few seconds and requiring
only the expansion of a few thousands of nodes in the search tree.
Problems with a up to $30$ and even $40$ objects are solved with
relative ease in the environment with three tables, but as expected become much harder when we have one single table, because
the objects clutter almost all available space, making it harder for the arm robot to move collision-free.
Indeed, the results show that the key parameter for scalability is \#c, which in a sense indicates
how cluttered the space is in the initial situation. When this number is not too high, as in the three-table environment,
our approach scales up with relative ease with the number of different specified goals.
Finally, preprocessing times scale up linearly with the number of objects, regardless of the number of goals,
thanks to the low-polynomial cost of the IW(2) pass on which the preprocessing is based, as detailed above.

\section{Discussion}
\label{sec:conclusion}


The presented work is closest to \cite{garrett2014ffrob,srivastava:2014}.
What distinguishes our approach is that combined task and motion planning problems 
are fully mapped into  classical AI planning problems  encoded in an expressive planning language.
Motion planners and collision checkers are used at compile  time but not at planning time. The approach is sound
(classical plans map into valid executable robot plans) and probabilistically complete
(with a sufficient number of configurations sampled, robot plans have a corresponding classical plan).
For the approach to be effective, three elements are essential. First, an expressive planning language
that supports functions and state constraints. Second,  a width-based planning algorithm that can
plan effectively for models expressed in  such a language without requiring the use of accurate but expensive heuristic estimators.
Third, a preprocessing stage that computes  the finite graphs of robot bases and arm configurations, 
the possible object configurations, and the  tables that allow us  to resolve procedural calls  into efficient table lookups. 
We have shown that the compilation process is efficient and independent of the number of objects, that the compiled problems are compact, and
that the planning algorithm can generate long plans  effectively.

For the experiments, we have considered the type of pick and place problems that have been used in recent work  \cite{garrett2014ffrob,srivastava:2014}.
For these problems, it is sufficient to  sample robot base configurations that are close to the physical tables,  and arm trajectories that can
pick up and place objects in the local space of the robot at a height that corresponds to the height of the tables. 
This part of the problem is not modeled explicitly in the  Functional STRIPS planning encodings, which implicitly assume
a finite graph of robot bases and one of robot arm configurations computed at preprocessing.
In the future, we want to represent this information explicitly in the planning encoding so that the
preprocessing stage can be fully general and automatic. This requires a general representation language
for CTMP problems so that the compilation will be a mapping between one formal language and another.
Unfortunately, there are no widely  accepted and shared formal models and languages for CTMP,
which makes it difficult to compare approaches empirically or  to organize `` CTMP competitions'',
that in the case of AI planning or SAT solving have been an essential ingredient for progress.
We believe that Functional STRIPS can actually serve  both roles:  as the basis for a general, integrated
representation language for CTMP problems and as a convenient  target language of the compilation representations.
This work is a first step towards this goal where we have shown that the compilation is indeed
feasible and effective  both representationally and computationally. 

\bibliographystyle{plainnat}
\bibliography{control}

\end{document}